\documentclass{article} % For LaTeX2e
\usepackage{nips15submit_e,times}
\usepackage{hyperref}
\usepackage{url}
\usepackage{graphicx}
\usepackage{amsmath}
\usepackage{natbib}

\title{Automatic Relevance Determination \\For Deep Generative Models}

\author{
Theofanis~Karaletsos\\
Computational Biology Program, Sloan Kettering Institute\\
1275 York Avenue, New York, USA  \\
\texttt{Theofanis.Karaletsos@ratschlab.org} \\
\And
Gunnar R\"atsch\\
Computational Biology Program, Sloan Kettering Institute  \\
1275 York Avenue, New York, USA \\
\texttt{gunnar.ratsch@ratschlab.org} \\
}

\nipsfinalcopy

\begin{document}

\maketitle

\begin{abstract} 
A recurring problem when building probabilistic latent variable models is regularization and model selection. For instance, the choice of the dimensionality of the latent space.
In the context of belief networks with latent variables, this problem can be adressed with Automatic Relevance Determination (ARD) employing Monte Carlo inference.
We present a variational inference approach to ARD for Deep Generative Models using doubly stochastic variational inference to provide fast and scalable learning. We show empirical results on a standard dataset illustrating the effects of contracting the latent space automatically. We show that the resulting latent representations are significantly more compact without loss of expressive power of the learned models.
\end{abstract} 

\section{Introduction}
In recent years, probabilistic treatments of deep learning architectures have spurred a renewed interest in the marriage of Graphical Models with the expressiveness of neural networks.
While Bayesian inference has traditionally been hard for nonlinear graphical models and based on sampling, recent efforts have focused more on deterministic approximate inference.
Variational inference methods for stochastic belief networks such as stochastic feedforward networks~\cite{tang2013learning}, the variational autoencoder~\cite{kingma2013auto} and related techniques~\cite{rezende2014stochastic,mnih2014neural,kingma2014semi,sohl2015deep} have been proposed and applied successfully in various contexts. 

Since these models are highly competitive as density estimators and bring the promise of being used as parts of larger graphical models, a fruitful area of research is the understanding and incorporation of structure on the latent spaces learned by such models. A key problem that remains unadressed in the context of these models is a principled criterion to determine the number of latent variables needed to model the data, especially in high-dimensional regimes.

While a trivial approach to selecting the dimensionality of the latent space is line-search over different settings with the marginal likelihood as a decision criterion, it is commonly of interest in probabilistic models to specify a large number for the available dimensionality and let the data evidence prune the useless dimensions in the posterior.

In this work we will study the use of doubly stochastic variational inference and the addition of a prior distribution over relevance-weights of latent dimensions to infer the effective dimensionality of the latent space in deep generative models and present experimental evidence of the effects on the inferred latent spaces.

\section{Background}

\subsection{Related Work}
An initial Bayesian treatment for regularization of Neural Networks was performed in seminal work by~\cite{mackay1995probable} and~\cite{neal1995bayesian}. They introduced a notion called Automatic Relevance Determination (ARD), which consists of the idea of using a prior distribution on generative weights attached to the latent variable which encourages the weights to be zero. Effectively, by integrating over such priors using Monte Carlo, settings for the variances of the prior can be inferred from data leading to pruning of unnecessary latent dimensions. ARD was also notably used in a variational treatment of the relevance vector machine~\cite{bishop2000variational} to infer a mask over the data features needed for a preditive model.
The idea of relevance determination for belief networks in combination with variational inference has also been explored before in~\cite{lawrence2002node}, but is different from our model.

Additionally, ARD is a key component of many Gaussian Process models~\cite{williams1996gaussian}, where it is used to select features of the input data to be passed through a covariance function.
A similarly inspired model to ours uses Bayesian Gaussian Process Latent Variable Models with ARD~\cite{damianou2012manifold} to learn rich latent spaces on multiple views, with the main difference being the use of Gaussian Processes instead of nonlinear parametric latent variable models.
Finally, a related inference method was presented in~\cite{titsias2014doubly}, where doubly stochastic variational inference was used for relevance determination on the input weights in logistic regression, but not in the context of deep generative models.

Apart from ARD, there is other recent work regularizing and manipulating deep generative models to add structure to the latent space. In~\cite{cheung2014discovering} a penalty term is introduced to force latent variables to decorrelate, while in~\cite{kohli_graphicsNetworks} known factors of variation are manipulated in controlled manual fashion during gradient descent to encourage learning of disentangled latent variables. Such approaches bring benefits in learning more interpretably shaped latent spaces and result in impressive performance gains.

\subsection{Stochastic Variational Bayes For Deep Generative Models}

We are presented with $N$ $D$-dimensional datapoints $x^i \in \mathcal{R}^{D}$.
We are interested in learning a probability distribution which captures the data well and thus maximizes $p({\bf x})$, for instance an embedding.
Let's assume a directed graphical model with latent variables $z^i$ corresponding to observed variables $x^i$. The latent variables can be drawn from any exponential family distribution $p(z)$, but simplifying cases for inference and learning exist for many continuous distributions.

Given a differentiable function $f$ parametrized by weights $\theta$, such as a multi-layer perceptron (MLP), we can write the model as follows:

\begin{align}
p(x ;\theta) = \int\limits_{z} p_{\theta}(x|z) p(z) dz
\end{align}
with 
\begin{align}
p_{\theta}(x|z) = f(x; z,\theta).
\end{align}

A good estimator for learning the parameters of such a model by assuming an approximate conditional posterior $q_{\phi}(z|x)$ was suggested in~\cite{kingma2013auto,rezende2014stochastic,mnih2014neural}, all of which can generally be understood as instances of doubly stochastic variational inference.
The estimator forms a variational lower bound~\cite{wainwright2008graphical,jordan1999introduction} to the marginal likelihood of the data. Thus, performing coordinate ascent with respect to variational parameters $\theta$ and $\phi$ corresponds to minimizing the divergence between the true posterior and the approximate one:

\begin{align}
\text{log}p_{\theta}(x^{(i)}) \geq \mathcal{L}(\theta,\phi;x^{(i)})=-\text{KL}(q_{\phi}(z)|| p_{\theta}(z)) + \text{E}_{q_{\phi}(z)} [ \text{log} p_{\theta}(x^{(i)}| z )].
\end{align}

We will denote this model with SGVB from now on. SGVB is quite robust to overfitting, as it optimizes a variational lower bound acting as a strong regularizer.

\section{Relevance Determination For Deep Generative Models}

We already pointed out the robustness of variational inference to overfitting. However, it is unclear whether this robustness extends to high-dimensional spaces. 
Regularization of complexity in deep unsupervised generative models can be performed in various ways: regularizing the previously deterministic weights, for instance by incorporating uncertainty using prior distributions on weights as suggested in the supplementary material of~\cite{kingma2013auto}, or inclusion of another stochastic variable which explicitly only controls the complexity of the prior in order to perform model selection.
Our contribution is concerned with the latter case, which is a simpler setting as it involves integrating over a relatively low-dimensional latent variable rather than all the weights in a complex network.
In order to address the problem of model selection for the dimensionality of the latent space, we re-introduce relevance determination as a sparsifying factor in the latent space. 

We can add relevance weights to the generative model by multiplying each latent variable dimension $z_d$ with a corresponding relevance weight $w_d$. We model $w$ using a Gaussian prior distribution $p(w)=\mathcal{N}(w;0,\Lambda)$, where $\Lambda = \text{diag}(\lambda_1,...,\lambda_D)$ denotes the diagonal covariance matrix.

The model then changes by incorporating two parents to variable $x$, namely variables $z$ and $w$:
\begin{align}
p(x ;\theta) = \int\limits_{w}\int\limits_{z} p_{\theta}(x|z,w) p(w) p(z) dz dw
\end{align}
with 
\begin{align}
p(x|z,w) = f(x; z\odot w, \theta)
\end{align}

\begin{figure}[ht]
\vskip 0.2in
\begin{center}
\centerline{\includegraphics[width=0.4\columnwidth]{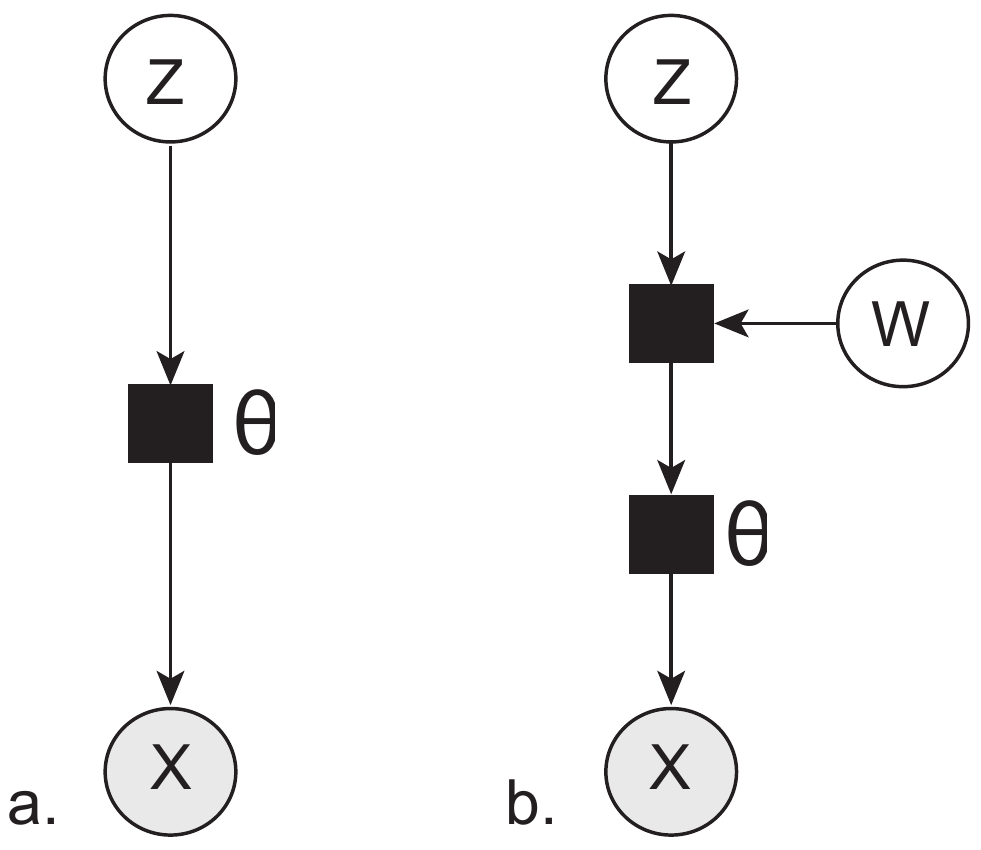}}
\caption{a. Graphical Model of SGVB. b. Graphical Model showing (G)SGVB-ARD}
\label{figure_graphical_model}
\end{center}
\vskip -0.2in
\end{figure}

In Figure~\ref{figure_graphical_model} we sketch how this corresponds to a graphical model with a more complicated factorial prior distribution than the one considered in~\cite{kingma2013auto}.

The particular structure of the relevance weights shows that they form a mask over the latent variable $z$, which is an input to the generative model $p_{\theta}$. 
The goal of adding this structure to the model is to facilitate the relevance weights to automatically reach distributions peaked around zero, in order to effectively prune dimensions of the input, which is a latent variable.
Apart from regularization and model selection the added variable also provides interpretability when using the model after being successfully inferred, by allowing the user to inspect the relevant latent variables directly.

Learning requires inferring optimal values for hyperparameters $\Lambda$ from the data. If $\Lambda$ becomes arbitrarily close to zero in any dimension, this dimension becomes meaningless to the generative model, as the prior has a fixed mean value of 0. We introduce an approximate posterior distribution over the weights $q(w; \tau)$ with $\tau_d=\{\mu_{\tau,d},\lambda_{\tau,d}\}$ for each dimension $d$. The goal now becomes to infer $q_{\tau}$ and use it to perform likelihood maximization on the prior parameters $\Lambda$ using the data evidence.
Such a procedure is called evidence maximization or empirical Bayes. An alternative would be to set a prior distribution over $\Lambda$, like a Gamma distribution, and infer the posterior values of $\Lambda$, at additional cost and complexity. In our case empirical Bayes corresponds to formulating and maximizing a lower bound to the marginal likelihood of the data with respect to the involved variational parameters $\theta$, $\phi$, $\tau$ and the hyperparameters $\Lambda$. The bound can be derived using rigorous variational inference principles~\cite{wainwright2008graphical,jordan1999introduction} and becomes as follows:

\begin{align}
\mathcal{L}(\theta,\phi,\tau,\Lambda;x^{(i)})=-\text{KL}(q_{\phi}(z)|| p_{\theta}(z)) -\text{KL}(q_{\tau}(w)|| p_{\theta}(w)) + \text{E}_{q_{\tau}(w)} \big[ \text{E}_{q_{\phi}(z)} [ \text{log} p_{\theta}(x^{(i)}| z,w )] \big].
\end{align}

As we can see, it involves nested expectations over approximate distributions $q_{\tau}$ and $q_{\phi}$, which are intractable to calculate analytically. We can apply the reparametrization trick from~\cite{kingma2013auto,rezende2014stochastic,titsias2014doubly} and use doubly stochastic variational inference to approximate the double expectation using just small numbers $N_w$, $N_z$ of samples $w^{l}$, $z^{k}$ from the approximate distributions $q_{\tau}(w)$, $q_{\phi}(z)$ respectively.

The bound then becomes:
\begin{align}
\label{SGVB-ARD-bound}
\mathcal{L}(\theta,\phi,\tau,\Lambda;x^{(i)})=-\text{KL}(q_{\phi}(z)|| p_{\theta}(z)) -\text{KL}(q_{\tau}(w)|| p_{\theta}(w)) +  \frac{1}{N_w}\sum \limits_{l}^{N_w}  \big[ \frac{1}{N_z}\sum \limits_{k}^{N_z} [ \text{log} p_{\theta}(x^{(i)}| z^{k},w^{l} )] \big].
\end{align}

and can be differentiated with respect to $\theta$, $\phi$, $\tau$ and $\Lambda$. We will denote this algorithm with SGVB-ARD henceforth.

\paragraph{Special Case: Gaussian Prior}
In the common case of a Gaussian prior on the latent space, we can exploit the model structure to simplify the variational objective. We notice the specific structure in the model $p(x)=\int \limits_{z,w} p_{\theta}(x^{(i)}| z,w )p(z)p(w)dz dw$ which involves a product of two normally distributed variables. We can replace the product with a joint Gaussian distribution over variable $m=z\odot w$ with $p(m)=  p(z)p(w) $ analytically and reach a closed form normal distribution for the bound in Equation~\ref{SGVB-ARD-bound}. Analytically, for the posterior distributions this becomes:

\begin{align}
\sigma_{\tau,\phi}^{2}=(\sigma_{\tau}^{-2}+\sigma_{\phi}^{-2})^{-1}\\
\mu_{\tau,\phi}=  \sigma_{\tau,\phi}^{2} \sigma_{\phi}^{-2} \mu_{\phi}+ \sigma_{\tau,\phi}^{2} \sigma_{\tau}^{-2} \mu_{\tau}\\
q_{\tau,\phi}(m)=\mathcal{N}(m;\mu_{\tau,\phi}, \sigma_{\tau,\phi}^{2})
\end{align} 

and similarly for the priors:
\begin{align}
\sigma_{m}^{2}=(\sigma_{w}^{-2}+\sigma_{z}^{-2})^{-1}\\
\mu_{m}= \sigma_{m}^{2} \sigma_{\phi}^{-2} \mu_{z}+ \sigma_{m}^{2} \sigma_{\phi}^{-2} \mu_{w}\\
p(m)=\mathcal{N}(m;\mu_{m},\sigma_{m^2})
\end{align} 
for each dimension. 

This reformulation can be helpful when writing the variational bound and applying the reparametrization trick since it reduces noise by just having a single expectation over which to evaluate the likelihood and a single regularizer in the latent space. We will denote this by GSGVB-ARD.

\section{Results}

In~\cite{kingma2013auto} it was noted that the performance of resulting models is roughly inependent of dimensionality and thus robust to picking a high dimensional latent space. In addition, in the more informal~\cite{kingmastochasticTalk} data is shown indicating that the squared weights as an indicator of importance per dimension of the topmost generative layer are automatically pruned when using SGVB and appear constant when changing the size of the latent space.
We perform a set of experiments to evaluate how robust SGVB is empirically to changes in size of latent space with respect to the implicit dimensionality of the latent space. We also compare to analogous experiments using SGVB-ARD and GSGVB-ARD.
We find that SGVB indeed prunes dimensions of $z$ automatically by pushing weights to zero over time during training and appears to use similar amounts of latent variables when changing dimensionality.
However, using ARD results in (sometimes significantly) more compact models in either cases at no cost to training-likelihood and reconstruction quality while typically boosting test-set likelihoods.
In all experiments we use minibatches of size 200 and found usage of rmsProp~\cite{graves2013generating} with momentum as an optimizer crucial for performance. We furthermore report the learning rates per experiment.
We initialize the $\Lambda$ parameters for the ARD weights to 1, such that ARD weights are free to fluctuate strongly a priori in all settings. In all cases we draw just one sample per datapoint in each iteration from the approximate variational distributions to perform coordinate ascend. We expect results to improve with using more samples.
We show detailed results in Table~\ref{logpx_table} and the details in the sections.

\begin{table}[ht]
\caption{Comparison of logP(x) of various models}
\label{logpx_table}
\vskip 0.15in
\begin{center}
\begin{small}
\begin{sc}
\begin{tabular}{lcccr}
\hline
%\abovespace \belowspace
Data set & SGVB & SGVB-ARD & GSGVB-ARD\\
\hline
%\abovespace
Frey Faces Train 400dh 50sh & 1162.36 & {\bf 1179.63}  & 1160.67 \\
Frey Faces Train 400dh 100sh & 975.06(1148.67) & {\bf 1209.54} & 1146.91 \\
Frey Faces Test 400dh 50 & 619.14 & {\bf 643.55}  & 583.87\\
Frey Faces Test 400dh 100sh & 547.17(627.35) & {\bf 667.42}   & 627.10\\
Frey Faces Train 200dh 50sh & 1104.88 & {\bf 1181.84} &1126.82\\
Frey Faces Train 200dh 100sh & 1155.09 & {\bf 1200.31} & 1145.12\\
Frey Faces Test 200dh 50sh & 637.47 & {\bf667.41} &  654.16 \\
Frey Faces Test 200dh 100sh & 653.06  & {\bf 686.95} & 673.30\\
\hline
\end{tabular}
\end{sc}
\end{small}
\end{center}
\vskip -0.1in
\end{table}

\begin{table}[ht]
\caption{Comparison of number retained latent variables}
\label{complexity_table}
\vskip 0.15in
\begin{center}
\begin{small}
\begin{sc}
\begin{tabular}{lcccr}
\hline
%\abovespace \belowspace
Data set & SGVB & SGVB-ARD & GSGVB-ARD\\
\hline
%\abovespace
Frey Faces 400dh 50sh &  27(30) & {\bf 9} & 19 (9)\\
Frey Faces 400dh 100sh & 26 & {\bf 9} & 21 (9)\\
Frey Faces 200dh 50sh & 21 & {\bf 9}  & 15 (9)\\
Frey Faces 200dh 100sh & 16 & {\bf 8} & 11 (9)\\
\hline
\end{tabular}
\end{sc}
\end{small}
\end{center}
\vskip -0.1in
\end{table}

\subsection{Frey Faces}

The Frey Faces dataset is a real-valued dataset showing faces in similar conditions. As it is not very large (under 2000 datapoints), it is prone to lead to overfitting when using hig-dimensional models in general.
We separate the model into 1600 training samples and 365 test samples.
We use 200 and 400 hidden rectified linear units, a Gaussian likelihood and latent variable with a zero mean and unit variance pior with 50 and 100 latent dimensions for a total of 4 experiments. We use a learning rate of 0.0001 in these experiments.

\begin{figure}[ht]
\vskip 0.2in
\begin{center}
\centerline{\includegraphics[width=0.95\columnwidth]{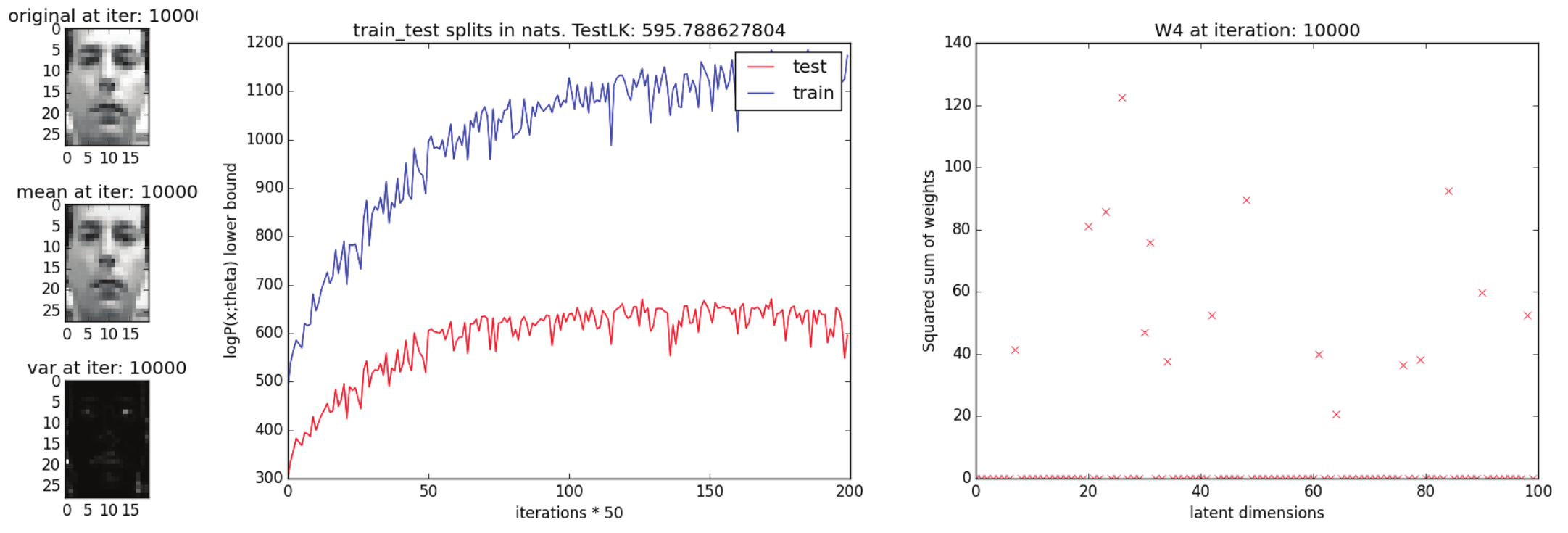}}
\caption{SGVB tile showing results for SGVB without ARD. To the right, an original image and the mean reconstruction of a model sample with variance over it. The middle shows a traceplot of the learning curve. The leftmost figure shows the squared sum of weights linked to each latent dimension, showing that SGVB prunes weights when needed.}
\label{FFnoARD}
\end{center}
\vskip -0.2in
\end{figure}

\paragraph{SGVB} We ran SGVB for the aforemetioned 4 experiments for 10000 iterations over the whole training-set. In this model, we observe that SGVB takes advantage of a large number of available latent dimensions, but eventually keeps between 16 and 27 dimensions respectively as input, see Table~\ref{complexity_table} and Figure~\ref{FFnoARD}.
Due to strong sampling noise when evaluating the likelihood we average over the test-scores of the last 100 iterations to reach the reported test-likelihoods of 637.47 nats for the 50-dimensional model and 653.06 nats for the 100-dimensional model with 200 deterministic hidden units.
The corresponding training scores are 1104.88 nats and 1155.09 nats, showing superiority of the model with more parameters on training data.

When using 400 hidden deterministic units, learning becomes unstable due to the paucity of data.
We report best achieved scores in brackets in addition to averages.
We observe similar trends as before, but the model uses 27/26 latent dimensions now instead of the 21 or 16 used with the better inferred model, indicating effects of overfitting.

\begin{figure}[ht]
\vskip 0.2in
\begin{center}
\centerline{\includegraphics[width=0.95\columnwidth]{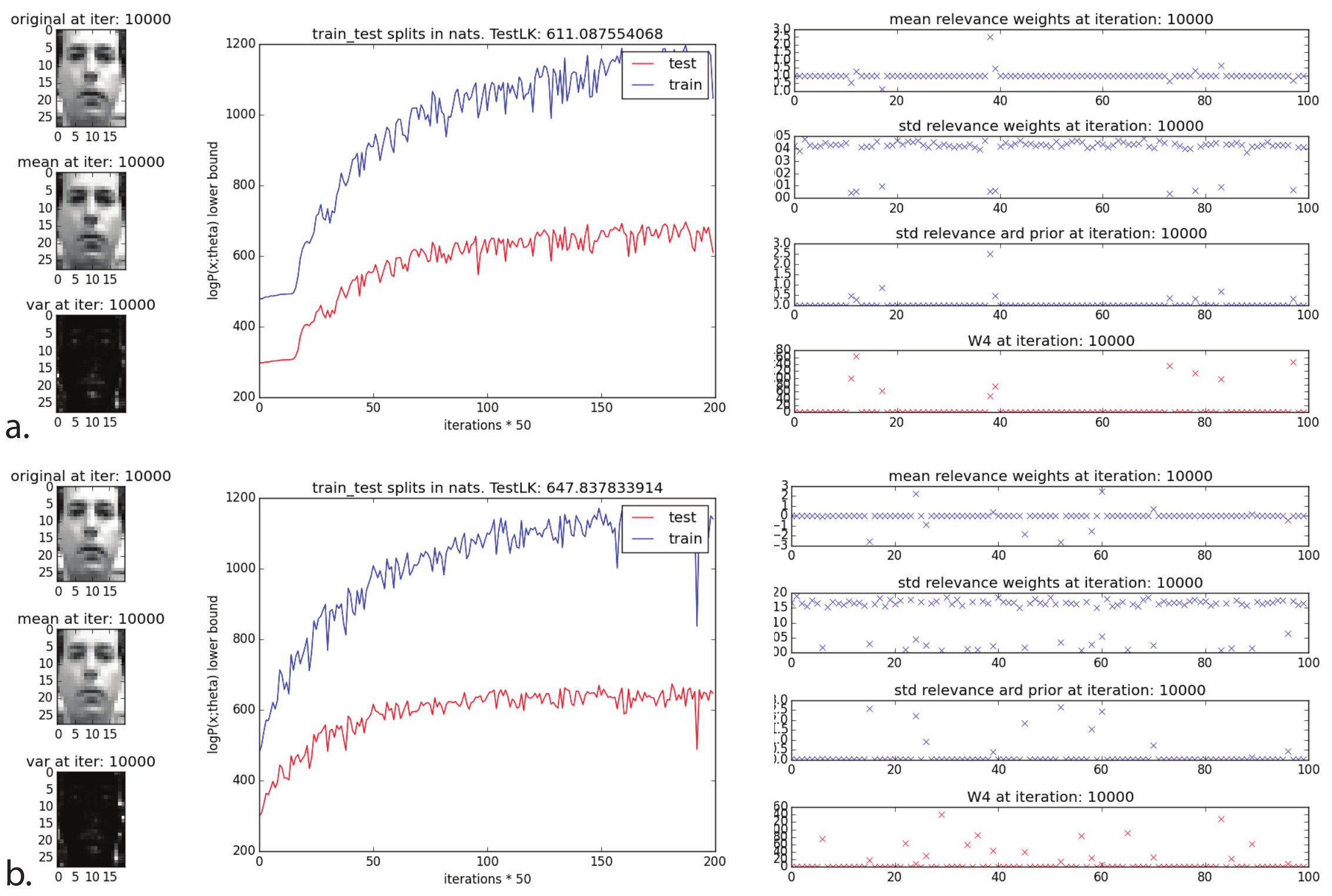}}
\caption{(G)SGVB-ARD tile (a) shows converged plots for SGVB-ARD with 100 hidden units. On the left, a reconstruction is shown at the end of training. The middle shows the learning traceplot. On the left, the relevance weight prior parameters and posterior parameters are shown.  (b) shows the same plots for GSGVB-ARD.}
\label{FFARD}
\end{center}
\vskip -0.2in
\end{figure} 

\paragraph{SGVB-ARD}We use the same settings for SGVB-ARD, to provide interpretable outcomes. 
During training we observe a very different behaviour of the lower bound of the training set. SGVB-ARD makes much slower progress than SGVB in the initial training phase. Monitoring reconstructions and relevance weights help us to understand this behaviour better.
Until about iteration 900, in both models with 50 and 100 latent variables, the variational bound makes slow progress while improving the reconstruction quality of the images slower than with SGVB. However, the uncertainty over the relevance weights is reduced systematically in these first 900 iterations. At about iteration 900, a phase switch occurs during which certain dimensions are pruned by setting relevance weights to small values and the model switches into a noisier training mode of optimizing reconstruction quality and exhibits behaviour closer to SGVB with fewer latent dimensions. After this initial phase SGVB-ARD makes faster progress during training. 
This can largely be attributed to usage of empirical Bayes, as the model initially deals with uncertainty in the relevance variables and initializes generative weights to a favourable regime. However, we observe that this uncertainty is gradually reduced over 1000 iterations instead of reaching a local minimum very quickly.
For all the described effects also see Figure~\ref{FFARD}.

We observe in Table~\ref{complexity_table} that SGVB-ARD retains 9 (once 8) latent variables stably across all settings, which is a significantly lower number than with the other approaches. While it can be argued that this may be a local minimum due to empirical Bayes, performance indicates that these models perform very well as they reach higher likelihoods in training and testing across all settings (see Table~\ref{logpx_table}) and show remarkable robustness to the data-poor setting when using 400 determainistic latent variables.

\paragraph{GSGVB-ARD}
GSGVB-ARD has the theoretical advantage of evaluating a single expectation rather than dealing with the stochastic approximations to two nested expectations. We expect this to result in 
This results in learning behaviour that resembles SGVB a lot more than SGVB-ARD, as no explicit phase-shift occurs during training and gradient noise is fairly stable. Inspecting the ARD weights also shows that they are not converged after the approximately 1000 iterations SGVB-ARD needs, prompting us to believe that learning is more evenly distributed over the learning process and thus more robust.

In terms of performance, we note that GSGVB-ARD also successfully prunes latent dimensions, produces qualitatively pleasing reconstructions and achieves high likelihoods. However, we note that it differs in performance from SGVB-ARD in two notable ways: First, GSGVB-ARD prunes less dimensions than SGVB-ARD but still more than SGVB. Interestingly it does this adaptively, as it retains more dimensions in the 400 hidden units setting than in the 200 hidden units setting.
Interestingly, close inspection of the relevance weights shows that many of the retained ones are close to 0 and the ones with significant posterior mass appear to be 9, which would indicate that similar models are retrieved as with SGVB-ARD but that training is not fully converged yet.
Second, in terms of likelihood, GSGVB-ARD performs on par with SGVB, but is outperformed by SGVB-ARD.However, learning was observed to be more stable to extremely noisy gradient steps than in the case of SGVB in the overfitting regime (400 units), which may be explained by the regularizing influence of the mask on the gradients. We also note that after 10000 iterations GSGVB-ARD was still improving, which may indicate that it may benefit from longer training times.

In summary, we observe that ARD benefits performance and compactifies learned models across the board. While SGVB-ARD exhibits 2 distinct phases in learning which may lead to fears of overfitting in more general settings, this issue is largely adressed in GSGVB-ARD due to analytical integration which smooths out the progress in learning across the two prior variables and leads to similar results.

\section{Discussion}

We have presented a method to perform fast, scalable feature selection in latent space using automatic relevance determination in nonlinear latent variable models trained with doubly stochastic variational inference. We observe that the application of ARD has strong contractive effects on the learned models in terms of variational compression and thus acts as a useful regularizer. In the context of unsupervised learning with disentangled and irreducible representations in mind, compactness is a  desireable property for generative models with the added benefit of speeding up computation. The reduction of the latent space propagates throughout the learned weights of the generative model as evidenced by the strong shrinkage of the top-most weights. This can have a profound effect on the learned representation and its expressiveness. We observe that variations in dimensionality frequently result in parsimonious final models using ARD, which suggests coherence in the learned models. However, further experimental evidence of the empirical behaviour of ARD in high-dimensional and data-poor settings is needed to elucidate its benefits. However, apart from the empirical success of the demonstrated method, access to a mask indicating relevance of particular dimensions also supports interpretability of learned models. This is a direct effect of using a factorial prior, which we believe will become more common in various contexts in deep generative models. It will be of interest for future work to study the effects of ARD on deeper and more structured generative models, especially if also used in intermediate layers, in terms of their semantic and subjective generative qualities and identifiability of the models in detail.
We finally note that a useful but potentially computationally costly alternative to ARD would be a fully Bayesian treatment of all weights in the network, as the network could learn to prune weights in the posterior when supported by data. The fact that just by using variational inference and without explicit shrinkage SGVB can learn to prune latent dimensions even without ARD further supports this view.

%\section*{Acknowledgments} 
%We wish to acknowledge David Balduzzi and Pedro Ortega for helpful comments and discussions.

\bibliography{ardVAE_arxiv}
\bibliographystyle{unsrt}

\end{document}